\documentclass[10pt]{article}
\usepackage{dsfont}
\usepackage[margin=3cm]{geometry}
\usepackage{amsmath}
\usepackage{algorithm}
\usepackage{algpseudocode}
\usepackage{authblk}
\usepackage{xcolor}
\usepackage{makecell}
\usepackage{multirow}
\usepackage{hyperref}
\usepackage{booktabs}
\usepackage{fancyvrb}
\usepackage{framed}
\usepackage{makecell}

\hypersetup{
    colorlinks=true,
    linkcolor=blue,
    filecolor=magenta,      
    urlcolor=blue,
    citecolor=blue,
    pdftitle={Overleaf Example},
    pdfpagemode=FullScreen,
    }
\usepackage[normalem]{ulem}
\usepackage[inline]{enumitem}
\usepackage{graphicx}
\graphicspath{ {images/} }
\usepackage{listings}
\usepackage{hhline}

\definecolor{codegreen}{rgb}{0,0.6,0}
\definecolor{codegray}{rgb}{0.5,0.5,0.5}
\definecolor{codepurple}{rgb}{0.58,0,0.82}
\definecolor{backcolour}{rgb}{0.95,0.95,0.92}

\lstdefinestyle{mystyle}{
    backgroundcolor=\color{backcolour},   
    commentstyle=\color{codegreen},
    keywordstyle=\color{magenta},
    numberstyle=\tiny\color{codegray},
    stringstyle=\color{codepurple},
    basicstyle=\ttfamily\footnotesize,
    breakatwhitespace=false,         
    breaklines=true,                 
    captionpos=b,                    
    keepspaces=true,                 
    numbers=left,                    
    numbersep=5pt,                  
    showspaces=false,                
    showstringspaces=false,
    showtabs=false,                  
    tabsize=2
}

\usepackage{authblk}
\title{From Barriers to Tactics: A Behavioral Science-Informed Agentic Workflow for Personalized Nutrition Coaching}
\author[1]{Eric Yang}
\author[1]{Tomas Garcia}
\author[1]{Hannah Williams}
\author[1]{Bhawesh Kumar}
\author[1]{Martin Ramé}
\author[1]{Eileen Rivera}
\author[1]{Yiran Ma}
\author[1]{Jonathan Amar}
\author[1]{Caricia Catalani}
\author[1]{Yugang Jia}
\affil[1]{Verily Life Sciences}
\date{}

\begin{document}

\maketitle
\begin{abstract}
Effective management of cardiometabolic conditions requires sustained positive nutrition habits, often hindered by complex and individualized barriers.
Direct human management is simply not scalable, while previous attempts aimed at automating nutrition coaching lack the personalization needed to address these diverse challenges.
This paper introduces a novel LLM-powered agentic workflow designed to provide personalized nutrition coaching by directly targeting and mitigating patient-specific barriers. Grounded in behavioral science principles, the workflow leverages a comprehensive mapping of nutrition-related barriers to corresponding evidence-based strategies. A specialized LLM agent intentionally probes for and identifies the root cause of a patient's dietary struggles. Subsequently, a separate LLM agent delivers tailored tactics designed to overcome those specific barriers with patient context.
We designed and validated our approach through a user study with individuals with cardiometabolic conditions, demonstrating the system's ability to accurately identify barriers and provide personalized guidance. Furthermore, we conducted a large-scale simulation study, grounding on real patient vignettes and expert-validated metrics, to evaluate the system's performance across a wide range of scenarios. Our findings demonstrate the potential of this LLM-powered agentic workflow to improve nutrition coaching by providing personalized, scalable, and behaviorally-informed interventions.
\end{abstract}
\section{Introduction}

The increasing prevalence of cardiometabolic diseases such as diabetes mellitus and hypertension poses a significant global health challenge \cite{National_Academies_of_Sciences_Engineering_and_Medicine2021-sl}.  Effective management of these conditions hinges on sustained lifestyle modifications, with nutrition playing a central role \cite{doi:10.1016/j.jacc.2023.12.023, Bermingham2024}. Digital health interventions, including mobile apps and online platforms, have emerged as readily accessible solutions. These platforms offer nutrition guidance, meal planning tools, and sometimes one-on-one interactions with human experts who aim to promote long-term adherence \cite{Majithia2020, Azelton2021, Lim2022}. However, traditional coaching models often face barriers like limited accessibility, high costs, and difficulties in scaling hyper-personalization.

The integration of large language models (LLMs) into digital health nutrition coaching presents an exciting opportunity to revolutionize personalized support, optimize intervention effectiveness, and improve access to care. LLMs possess the ability to engage in natural language conversations, enabling them to provide dynamic interactions that go beyond traditional static information delivery. Notably, their conversational ability allows LLMs to clarify user intent, answer nutrition-related questions with contextual nuance, provide tailored recommendations that align with individual preferences and needs, and adapt their approach based on user feedback \cite{tu2024conversational, peng2023study, singhal2022large}. This personalized conversational approach, combined with the accessibility and scalability of digital platforms, holds immense promise for transforming nutrition coaching and improving the health of individuals around the world. In addition, recent research has showcased LLMs' potential in gaining user acceptance and delivering accurate domain-specific knowledge, specifically in nutrition applications addressing cardiometabolic conditions \cite{https://doi.org/10.48550/arxiv.2312.04059, Ong2024, 10.1145/3613904.3641924, Sun2023}.

However, simply delivering nutrition information through general foundation models is unlikely to achieve sustained behavior change. Leveraging the principles of behavioral science is critical for designing sustainable interventions that address the physical, psychological, and social factors influencing dietary choices. Frameworks such as the capability-opportunity-motivation-behavior (COM-B) model are instrumental in understanding the multi-dimensional factors that contribute to current behavioral patterns \cite{Michie2011}. Once the factors preventing behavioral change are better understood, additional frameworks such as the Behavioral Change Wheel (BCW), the Behavioral Change Taxonomy of 93 hierarchically clustered techniques (BCT) and the Easy, Attractive, Social and Timely (EAST) framework can provide the right approach and tactics on how to promote change \cite{Michie2013-xg, East, Michie2011}. These frameworks offer insights into designing interventions that leverage psychological principles to enhance engagement, simplify behavior, make actions rewarding, and trigger positive associations. Combining these complementary frameworks, addressing both the root causes and offering solutions, is crucial to realize LLMs' potential in delivering sustainable behavioral change. 

While recent research has explored the potential of LLMs for digital coaching in health behavior change, several opportunities remain to enhance the integration of behavioral science principles. Many interventions do not comprehensively quantify behavioral barriers and often rely on high-level motivational strategies rather than providing targeted, actionable behavior change tactics \cite{Jorke2024, Hegde2024, Meywirth2024, Kumar2024, Dao2024, Bak2024}. For instance, some approaches utilize LLMs to deliver advice primarily through broad motivational interviewing techniques or empathetic tones. However, these methods may benefit from a more explicit focus on quantifying barriers from a behavioral science perspective and offering specific tactics tailored to individual needs. Similarly, other studies have incorporated frameworks like COM-B to identify barriers, but reliance on single conversational turns for barrier classification may limit the depth of assessment. A more iterative probing process could enhance the identification and understanding of these barriers. Overall, there is significant potential for a more nuanced integration of behavioral science expertise in the design and development of LLM-powered digital coaching interventions to support more effective and sustainable behavior change.

\subsection{Contributions}

\begin{figure}[t]
\caption{\textbf{Summary of Contributions.} (A) We introduce a multi-agent AI Coach workflow that is infused with behavioral science principles to unblock patients towards their nutrition goals. The workflow consists of two core agents, the barrier identification agent and the strategy execution agent. (B) To assess the performance of our workflow at scale, we developed patient simulators that portray validated nutrition vignettes drawn from real cardiometabolic patient profiles. (C) Real cardiometabolic patients and patient simulators interacted with our multi-agent AI coach workflow. The conversation experiences are evaluated via survey, expert annotation of dialogues, and are additionally compared to a baseline LLM via auto-evaluation.}
\centering
\includegraphics[scale=.31]{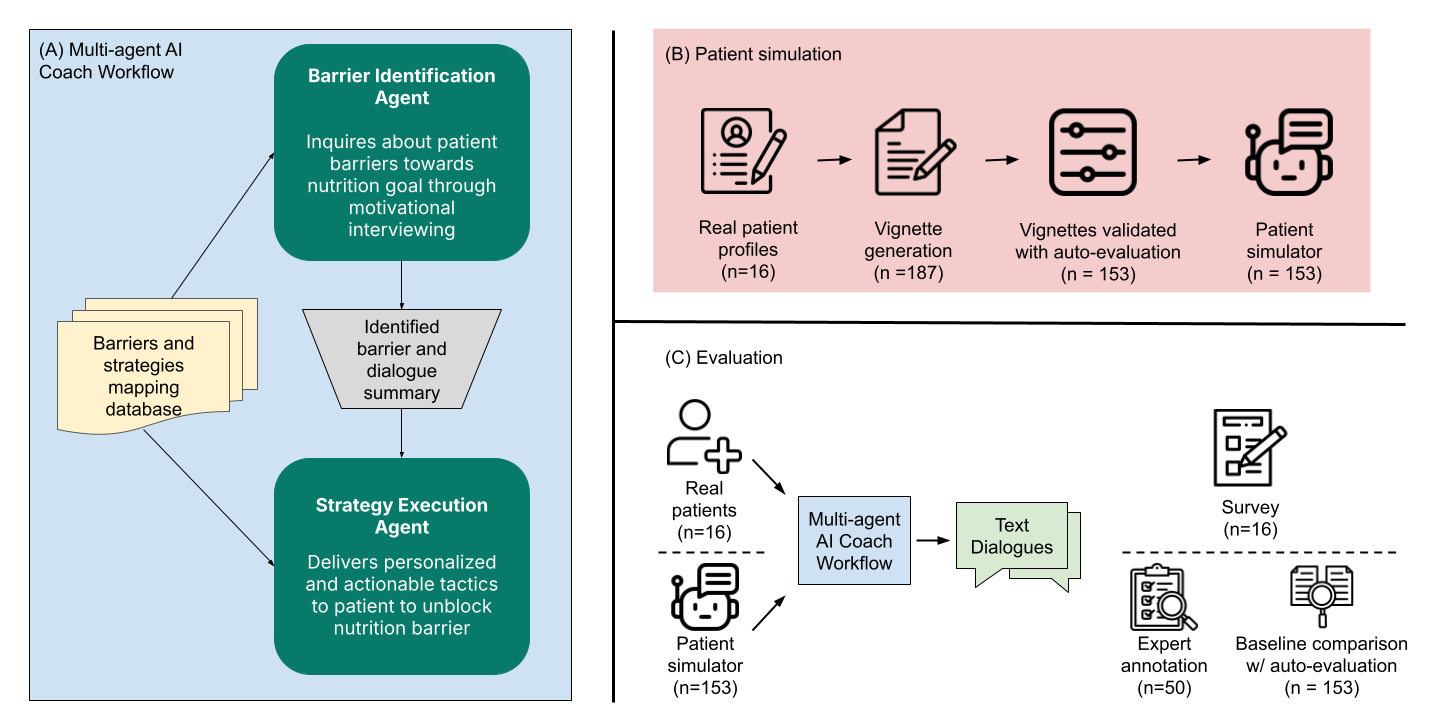}
\label{fig:overall-contributions}
\end{figure}

We introduce a novel approach in leveraging LLM-powered nutrition coaching for cardiometabolic condition management by advancing the integration of behavioral science principles and developing a comprehensive, scalable, and expert and user validated framework. Our research directly addresses current limitations in the field by creating a multi-agent conversational workflow that is powered by a deep understanding of nutrition-related barriers and corresponding strategies that directly mitigates the barriers. Through motivational probing, our approach directly identifies the root causes of nutrition behavior rather than addressing the surface-level symptoms. It fosters a personalized coaching experience, moving beyond high-level motivational techniques and offering targeted tactics supported by behavioral science research. Our approach also enables a learning system, where barrier-strategy mappings can be tailored to specific individual context, habits, and tactic adoption. We summarize our contributions below:

\begin{enumerate}
    \item \textbf{Comprehensive Barrier Identification and Strategy Mapping:} We conducted a user research study to identify common nutrition-related barriers experienced by cardiometabolic patients. In addition, we performed comprehensive literature review of academic research papers and internal reports. We then synthesized these barriers into an overarching set of main barriers encompassing all aspects of nutrition goal achievement. Furthermore, we mapped these barriers to a comprehensive set of strategies and behavioral science tactics, enabling a tailored approach to addressing barrier-specific challenges. This goes beyond existing research by providing a more holistic and nuanced understanding of nutrition barriers and their corresponding solutions, paving the way for targeted interventions.
    \item 
    \textbf{Multi-Agent Workflow for Personalized Coaching:} Building on our comprehensive barrier and strategy mapping and insights from user research, we designed a multi-agent workflow where a specialist LLM agent is tasked with probing and classifying barriers through conversation, while another specialist LLM agent carries out strategies and offer specific tactics. This approach improves upon existing single-agent systems, fostering more convenient performance optimization in different stages of behavioral change delivery.
    \item 
    \textbf{Real-World Validation with Cardiometabolic Populations:} To ensure practical application and effectiveness, we validated our workflow with participants representative of the cardiometabolic populations we aim to serve. This direct hands-on validation provides evidence for the efficacy of our system in addressing real-world nutrition challenges, enhancing the credibility and impact of our research. The real-world validation sets our work apart from purely simulation frameworks.
    \item 
    \textbf{Benchmarking and Expert Annotation:} Along with behavioral science experts, we propose a granular benchmark for evaluating the performance of our LLM agents in various stages, including barrier identification, tactic delivery, and overall conversational attributes.  Expert annotations on these metrics provide a rigorous evaluation of our system's performance. 
    \item \textbf{Simulation-Based Evaluation at Scale}: We curated real patient vignettes from our user study to generate large-scale simulated conversations of various realistic barrier situations. This data is then evaluated by LLM auto-evaluators, allowing for a scalable and accurate assessment of our system's performance across a diverse range of scenarios. This simulation-based evaluation approach provides a scalable method for evaluating the generalizability of our LLM-powered coaching system, setting a standard for evaluating the performance of similar AI-driven interventions.
\end{enumerate}
We structure the remainder of the paper as follows. Section \ref{sec:uxr-insights} outlines the user research study conducted to help inform the coaching workflow design. In Section \ref{sec:workflow-design}, we describe the architecture of the behavioral science-informed agentic workflow. Section \ref{sec:uxr-eval} presents the user research study conducted to evaluate the workflow with real users. In Section \ref{sec:simulation-study}, we outline the simulation study designed to test the workflow's effectiveness in various scenarios. Section \ref{sec:eval-results} reports the results with expert and auto-evaluation of the workflow. Section \ref{sec:discussion} provides a discussion of the findings and their implications for future research and practice.

\section{User Research Study: Informing Workflow Design}
\label{sec:uxr-insights}
Conducting a user research study is crucial for designing a coaching workflow with the specific needs of individuals with cardiometabolic conditions in mind. By directly involving the target population, we aimed to gain a deeper understanding of realistic nutrition challenges and preferences, ensuring that the system is designed to provide relevant, practical and effective guidance. 

This research study aimed to uncover key user insights necessary for building a patient-centered AI coaching workflow. Specifically, we looked to answer two main research questions 1) What user motivations, characteristics, and challenges must our workflow understand and adapt to in order to be effective? 2) How can an AI coach's character and conversational patterns inspire trust and engagement while delivering behavioral science strategies?
\subsection{Methods Overview}
Sixteen participants were recruited through a purposeful sampling strategy, selecting for diversity in demographics, United States geographic regions, health conditions, attitudes toward health care and technology, and recent cardiometabolic diagnosis. Participants were recruited for sessions across a 12 week period, and were compensated at the rate of \$100 per hour. 

Each participant took part in one-on-one, semi-structured qualitative interviews led by the research team. These interviews focused on participants' medical histories, the barriers they encountered when working toward their health goals, and their responses to different conversational approaches. Participants shared their experiences and reflected on the specific challenges they faced in managing their health, while also discussing strategies they had found useful. Additionally, to address the second research objective, participants also engaged in interactive exercises where they conversed with two types of LLM agents: a supportive agent and an assertive agent. The supportive agent facilitated conversation by treating the user as the expert on their body and experience. This agent encouraged self-reflection and demonstrated high levels of compassion, curiosity, and affirmation. Its tone was plain-spoken, easygoing, and patient, kindly taking direction from the user. In contrast, the assertive agent adopted a directive approach, positioning itself as the resident expert with high energy, authoritative knowledge, and a strategic mindset. This agent aimed to empower the user by being assertive and eager to motivate change. Both agents are powered by Gemini-1.5 Pro, a commercially available model known for its conversational and reasoning capabilities \cite{team2023gemini}. Their character and tone were guided by specific phrasing and conversational styles in single prompts. 

Analysts used a modified-grounded approach to qualitative analysis, using prior behavior change theories and user experience expertise in the data analysis process. This approach allowed for a focused exploration of specific aspects of AI desirability and behavior change motivation, while grounding observations on the data itself.
\subsection{Research Insights}
The user research study revealed crucial insights into the barriers and motivations that should be integrated into the design of the AI coaching workflow. Participants shared detailed accounts of their physical, psychological, and social challenges in achieving their health goals. Common themes that emerged included balancing competing priorities, dealing with physical limitations, and struggling with low self-efficacy. For example, participants frequently expressed feeling overwhelmed by the demands of daily lives, which made it difficult to sustain motivation to work on their health over time. 

Additionally, participants highlighted the importance of feeling understood and empowered by a coach. A preference emerged for the supportive agent, as most participants felt more comfortable and engaged with its empathetic approach. They noted that the supportive agent fostered trust and provided a space for self-reflection. While a few participants appreciated the assertive agent’s high-energy style, the majority found it to be less aligned with their desire for a collaborative and self-empowering experience. This feedback was instrumental in refining the AI coaching workflow, which ultimately prioritized the compassionate, user-driven style of the supportive agent to promote long-term engagement and behavior change. 

These insights directly informed the design of our AI coaching workflow. We recognized the importance of understanding the root causes of individuals' behavioral barriers before suggesting practical, incremental steps to help them overcome their challenges. This ensures that the solutions offered are not only relevant but also actionable and aligned with the user’s personal context. Additionally, the user research allowed us to ground the relevance of the barriers and strategies that are provided to the target cardiometabolic population we aim to serve. Finally, the study highlighted the importance of a conversational style that fosters trust and engagement, shaping the development of a communication approach that feels supportive and empowering.

\section{Behavioral Science Agentic Workflow Design}
\label{sec:workflow-design}
\subsection{Curation of Barriers and Behavioral Science Strategies}
Recognizing that the barriers described by the participants in the user study may not be all encompassing, we additionally researched the challenges faced by people with cardiometabolic conditions in the literature and marketplace. We analyzed numerous research papers and reports, covering areas like nutrition, medication adherence, exercise, and goal setting. This helped us identify over 100 total barriers individuals experience.
Next, we used affinity mapping to organize these barriers, grouping them based on common themes and patterns. By prioritizing the most frequent and impactful challenges, we narrowed our focus to 28 key barriers.

To uncover effective strategies and tactics for addressing these barriers, we conducted a comprehensive review of existing behavioral science frameworks and developed a solutions repository.
First, we examined established frameworks, seeking strategies and tactics with proven efficacy in overcoming similar barriers within the health domain. Our review encompassed well-known models like the COM-B model, the BCT Taxonomy, the EAST framework and other behavioral change models \cite{Michie2011, Michie2013-xg, East}.
Following this extensive mapping exercise, we curated a final selection of repository containing over 50 strategies and 100 tactics that are mapped to the 28 barriers. This repository comprehensively links each identified barrier to a range of potential solutions, offering guidance on optimal implementation to maximize impact. While we recognize the value of popular frameworks and pre-defined sequences for intervention design, we deliberately chose not to explicitly link our workflow to any singular pathways at this stage. This decision allows for greater flexibility and innovation as we explore the full potential of our future capabilities, ensuring we're not constrained by any pre-determined intervention pathways. Few examples of barriers, strategies and tactics mapping are shown in Table \ref{tab:best-mappaings-examples}.

\begin{table}[ht!]
\centering
\caption{Examples of Barriers, Strategies and Tactics Mapping}
\label{tab:best-mappaings-examples}
\begin{tabular}{ | m{4.75cm} | m{4.75cm}| m{4.75cm} |} 
    \hline
    \textbf{Barriers} & \textbf{Strategies} & \textbf{Tactics} \\ 
    \hline
    Decision fatigue: the mental exhaustion from making to many choices can lead to poor, sub-optimal future decisions. E.g. \textit{``As a working parent I have to deal with so many things that I don't want to think about cooking when I get home."}  \cite{Pignatiello2018} &  Heuristics: Mental heuristics are cognitive shortcuts our brains use to simplify complex situations and make quick decisions. \cite{Moorhouse2020} &  (i) Rules of thumb: Offer the user a set of actionable principles to help them automate tasks. E.g. \textit{``Always, fill one third of your plate with lean protein."} (ii) Default: Encourage the user to pick an option and use it as a default to save time and mental effort. E.g. \textit{``Let's set Tuesday as a kale salad day!"} \cite{Moorhouse2020}  \\ 
    \hline
    Present bias:  It's the user tendency to overvalue immediate rewards over future, larger rewards. E.g. \textit{``It's the weekend! let's start with diet next week! "} \cite{Chakraborty2021} & Future self. Future self is a vivid and emotional connection with a future version of ourselves that affect our intention to engage in a future behavior \cite{Hershfield2011}. & (i) Mental rehearsal of successful performance: Encourage the user to practice visualizing themselves successfully performing the desired behavior in realistic scenarios. For example, suggest they vividly imagine eating greens and feeling light to continue with the day. \cite{Michie2013-xg}\\
    \hline
\end{tabular}
\end{table}

\subsection{Multi-Agent Workflow}
This section details the architecture of the multi-agent LLM coaching workflow, aiming to provide tailored nutrition guidance to cardiometabolic patients by identifying and addressing their unique barriers to achieving their nutrition goals. As shown in Figure \ref{fig:overall-contributions}, the system operates through two core agents, seamlessly integrated to offer a unified user experience: the Barrier Identification Agent and the Strategy Execution Agent. Both agents are powered by Gemini-1.5 Pro, and are prompted to converse in the supportive manner preferred by participants in the user study.

\subsubsection{Barrier Identification Agent}
The Barrier Identification Agent initiates the conversation by inquiring about the user's current nutrition goal and progress. It then employs motivational interviewing techniques to explore the specifics of the user's nutrition habits, with a focus on identifying the most prominent barrier that may be hindering progress toward their stated goal. The agent is equipped with a predefined taxonomy of 28 barrier concepts, provided within the prompt alongside detailed descriptions and examples. Through iterative dialogue, the agent analyzes the user's responses to classify the identified struggles into one of the barrier concepts. While a user may exhibit multiple barrier concepts, the agent is instructed to prioritize and focus on the most prominent one displayed by the user. Once the agent determines that sufficient information has been gathered for barrier classification, it generates an internal summary of the conversation and the identified barrier concept, which is then relayed to the Strategy Execution Agent. The prompt used by the Barrier Identification Agent can be found in Appendix \ref{barrier_id_prompt}.

\subsubsection{Strategy Execution Agent}
The Strategy Execution Agent receives the conversation summary and identified barrier concept from the Barrier Identification Agent. Based on this input, it first retrieves corresponding tactics and execution sequences from a predefined table. Each barrier concept is mapped to a set of potential tactics, each with examples and an associated execution sequence outlining mandatory and optional tactics, as well as their prioritization. The agent then engages in further conversation with the user, drawing upon the preceding conversation summary and the prescribed tactics. For the optional tactics, the agent dynamically adapts its offerings based on user responses and perceived acceptability and effectiveness of the previously deployed tactics. The conversation concludes when the agent determines the user has been sufficiently equipped with the necessary tools to overcome their identified barrier and progress toward their nutrition goal. The prompt used by the Strategy Execution Agent can be found in Appendix \ref{strategy_exec_prompt}.

\subsubsection{Agent Orchestration}
While the back-end functionality relies on two core distinct agents, the user interacts with the system as a single, continuous AI-powered nutrition coach. This seamless transition between barrier identification and strategy execution is facilitated by the orchestration of information exchange between agents. Crucially, conversation summaries and key outputs, including identified barriers and nutrition stories, are explicitly transferred between the agents, ensuring the preservation of all necessary context while excluding irrelevant banter. This modular design allows each agent to specialize in its respective task, enabling more focused model improvements for both barrier identification and strategy execution. An example of the agent orchestration can be found in Appendix \ref{sample_convo}.

\section{User Impressions of Workflow}
\label{sec:uxr-eval}
\subsection{Research Objectives}
Having implemented the AI coaching workflow grounded on the insights from the initial user research, we conducted further sessions to evaluate users' impressions of the workflow through hands-on interactions. This phase aimed to test the real-world applicability and accuracy of the system's barrier identification and strategy execution capabilities, providing a critical assessment of how well the workflow translates theoretical principles into practical outcomes. Engaging with users also helps us uncover any usability or communication insights that may not have been apparent in agentic workflow and prompt development. Key questions guiding this evaluation were:
\begin{itemize}
\item \textbf{Effectiveness of Behavioral Science-Informed Workflow}: How do users who are managing health concerns respond to our workflow infused with behavior science frameworks? How effectively does our workflow help users identify obstacles that prevent them from being healthier? Are the strategies and tactics offered by our workflow easy for users to put into action? How does our workflow impact users’ motivation and confidence to make positive changes in their health?
\item \textbf{Building Trust and Engagement through AI Interactions}: Does our AI coach's character and personality inspire trust and engagement? To what degree do users feel supported by our coaching workflow?

\end{itemize}
\subsection{Methods Overview}
In this session, a subset of 6 participants were selected for hands-on conversational interaction with our LLM coaching workflow. The subset was selected on the basis of diversity with respect to AI comfort levels and stated motivation levels. The participants were primed to think of challenges they might face achieving a nutrition goal they had set, and to channel that while interacting with our coaching workflow. After the experience, participants were asked to reflect on their interactions through qualitative interviews and complete a survey. Analysts used a modified-grounded approach to qualitative analysis.
\subsection{Research Results}
\textit{Effectiveness of Behavioral Science-Informed Workflow}. The AI coaching workflow, grounded in behavioral science principles, was found to be effective in unblocking barriers users have toward healthier habits. First, the AI coach was successful in helping users identify specific barriers to their health. 5 out of 6 participants agreed that the assistant helped them recognize obstacles that prevented them from being healthier. Participants also indicated that their interactions with the AI were informative. When asked whether they learned something new about their health habits, 4 out of 6 participants strongly agreed and the remaining somewhat agreed with the sentiment. As a participant reflected, the agent’s approach of leading with additional questions and building from user experiences enabled “more refined, bite-sized” understanding, enhancing quality of the nutrition motivations and barriers insights collected. Furthermore, participants responded positively to the strategies and tactics provided by the AI coach. All participants strongly agreed that the AI coach’s advice felt personalized to their situation, reflecting its ability to adapt to users' unique needs. Notably, all participants agreed that the AI coach's advice was easy to put into action, highlighting its practicality. Participants appreciated the focus on manageable, small changes, with one stating, “Making small changes makes a big difference," highlighting the perceived effectiveness of the behavior change strategies recommended by the AI coach. The AI coach's ability to offer novel suggestions tailored to individual preferences was highlighted by participants, such as when one participant appreciated the recommendation to swap quinoa porridge for oatmeal, and another found value in being advised on alternatives to high-sugar protein drinks. The impact of these strategies on user motivation was significant, with all participants agreeing that the AI coach increased their confidence to make positive changes in their health. Each participant also reported feeling more motivated after their conversations due to excitement for having a plan to implement manageable, novel, small changes.

\textit{Building Engagement through AI Interactions}. Overall, the AI coaching workflow fostered relationship building and engagement through their conversational style and empathetic personality. All participants strongly agreed that they felt supported by the AI coach, and were interested in having future conversations about their health, indicating a high level of engagement. One participant noted “I could sit there and talk to that thing all day.”  Furthermore, participants frequently described the AI as having a “friendly” and “human-like” demeanor, with one participant mentioning, “It did feel like you'd spoken with somebody, like you had an actual conversation with somebody." As one participant noted, the AI coach “seemed to be friendly, kind of human-like,” creating a sense of connection and understanding that facilitated deeper discussions about their nutrition habits. The AI coach’s ability to create a supportive environment was evident, with all participants agreeing that they felt comfortable discussing their health issues with them. This sense of comfort was further supported by the finding that the great majority of participants strongly agreed that their conversations were engaging. The AI coach's empathetic and person-centered communication style not only facilitated open dialogue but also encouraged continued user interaction, which is crucial for sustained engagement and positive health outcomes. The full survey results can be found in Appendix \ref{uxr_survey}.

\section{Simulation study}
\label{sec:simulation-study}
\subsection{Patient Vignettes}
To assess the performance of our multi-agent LLM coaching workflow at a larger scale, we conducted a simulation study with the high level workflow illustrated in Figure \ref{fig:overall-contributions}. Leveraging patient profiles curated from the user research study that contained patient lifestyle context and medical history, we identified the prominent nutrition barriers for each profile, aligning them with our predefined taxonomy of 28 barrier concepts. This involved a manual process of translating patients' self-reported nutrition struggles into corresponding barrier concepts. Subsequently, for each identified barrier within each profile, we crafted a detailed patient nutrition vignette paragraph with a separate base Gemini model. These vignettes aimed to vividly depict the specific manifestation of that barrier within the context of the individual's profile. Providing pre-written nutrition vignettes allowed us to:
\begin{itemize}
        \item Focus the Patient Simulator: Our approach enabled the downstream patient simulator to concentrate on portraying the specific barrier and engaging in a meaningful conversation,  without the need to generate complex nutrition narratives dynamically.
    \item Enhance Control and Consistency: We enabled greater control over the simulated conversations, minimizing the risk of inconsistencies or incoherent narratives that might arise from on-the-fly story generation.
    \item Isolate Barrier Identification Impact: Our approach allowed us to isolate and assess the system's ability to accurately identify and address specific barriers, independent of other confounding factors within a patient's profile.
\end{itemize}

In total, we simulated a wide range of scenarios across diverse patient profiles, nutrition goals, and barriers, resulting in 187 total simulated vignettes. We provide a sample vignette generated from a barrier concept for a patient profile in Appendix \ref{sample_vignette}.

To ensure the quality and relevance of the generated vignettes with respect to the simulated behavioral barrier, we leveraged OpenAI's GPT-4o (\texttt{gpt-4o-2024-08-06}) model in a LLM-as-judges framework, also known as auto-evaluation \cite{openai_gpt4o}. The GPT-4o model was intentionally chosen as the vignettes were generated by a Gemini model. The complete prompt used for the auto-evaluator is included in Appendix \ref{eval_vignette_prompt}. The auto-evaluator received two inputs: the target barrier to be simulated, and the generated patient vignette. It then evaluated the vignette across four dimensions:

\begin{itemize}
    \item \textbf{Evidence (High/Medium/Low):}  The extent to which the vignette provides clear indications that the patient's behavior or thoughts are influenced by the target barrier.
    \item \textbf{Realism (High/Medium/Low):}  The plausibility and believability of the depiction of the target barrier, reflecting how it might manifest in a real person's life.
    \item \textbf{Completeness (High/Medium/Low):}  The sufficiency of detail provided in the vignette to fully understand the impact of the target barrier on the patient's ability to achieve their nutrition goal.
    \item \textbf{Leakage (Yes/No):}  Whether the vignette directly mentions the technical term of the target barrier. It is important that vignette does not explicitly contain the technical term, which may skew the barrier identification task down the line.
\end{itemize}

To calibrate the auto-evaluator's performance, we conducted an adversarial analysis. We evaluated the auto-evaluator on a patient vignette paired with a randomly chosen behavioral barrier different from the one simulated in the vignette. This mismatch was designed to assess the auto-evaluator's sensitivity to inconsistencies between the intended barrier and the generated narrative. We hypothesized that scores for these adversarial cases would be significantly lower, reflecting the lack of alignment between the vignette and the incorrect barrier. The results of this analysis are presented in Table \ref{tab:auto-eval-results}.

\begin{table}[ht!]
\centering
\caption{Auto-Evaluator Performance on Matched and Mismatched Vignettes}
\label{tab:auto-eval-results}
\begin{tabular}{l|ccc|ccc}
\toprule
 & \multicolumn{3}{c|}{\textbf{Matched Vignette}} & \multicolumn{3}{c}{\textbf{Mismatched Vignette}} \\
  & \multicolumn{3}{c|}{(n=187)} & \multicolumn{3}{c}{(n=187)} \\
\midrule
\textbf{Dimension} & \textbf{High} & \textbf{Medium} & \textbf{Low} & \textbf{High} & \textbf{Medium} & \textbf{Low} \\
\midrule
Evidence & 184 & 3 & 0 & 36 & 32 & 119 \\
Realism & 187 & 0 & 0 & 84 & 94 & 9 \\
Completeness & 156 & 31 & 0 & 19 & 46 & 122 \\
\bottomrule
\end{tabular}
\end{table}

Of the 187 simulated vignettes, 153 of them received high marks across evidence, realism, and completeness dimensions by the auto-evaluator. These higher quality vignettes were selected for downstream conversation simulation. As hypothesized, the auto-evaluator assigned significantly lower scores to the mismatched vignettes across all dimensions. This indicates the auto-evaluator's ability to discern between accurately and inaccurately represented barriers, supporting its validity for assessing vignette quality. In addition, there was no barrier term concept leakage across all cases. Our vignette generation and validation processes provided a robust foundation for our simulation study, ensuring realistic and diverse scenarios for evaluating the effectiveness of our multi-agent LLM coaching workflow.

\subsection{Conversation Simulation}
Having established a robust set of patient vignettes, we proceeded to simulate dialogues between a Gemini-powered patient simulator and our LLM coaching workflow for each of the 153 higher quality vignettes. The patient simulator was provided with two key inputs: the generated vignette from the corresponding patient profile and barrier, and the communication style observed for that patient during the user research study. Notably, the patient simulator's nutrition barriers are only portrayed via the vignette without explicit mentioning of any technical behavioral science barrier terms. The patient simulator, guided by these inputs, engaged in a conversation with the AI coach. The dialogue continued until the AI coach determined, based on its internal logic and the patient's responses, that the patient had reached a higher level of preparedness to address their nutrition-related challenges. This endpoint represented a complete coaching interaction within the simulation framework. The prompt used by the patient simulator can be found in Appendix \ref{patient_simul_prompt} and an example simulated conversation can be found in Appendix \ref{sample_convo}.

\subsection{Expert Assessment}
Evaluating the quality and effectiveness of the simulated coaching conversations was a crucial step in assessing the overall performance of our multi-agent LLM workflow. We developed an evaluation rubric encompassing five key dimensions: barrier identification accuracy, tactic comprehensiveness, tactic personalization, tactic actionability, and conversation empathy. These dimensions were chosen to reflect the core competencies required for effective digital coaching in the context of motivating nutrition behavioral change.

\begin{itemize}
\item \textbf{Barrier Identification Accuracy:} Given the list of 28 nutrition barrier concepts, the AI coach identified the correct patient barrier. Accurate identification of the patient's primary barrier is paramount for delivering tailored and effective interventions. Rating Options: “Yes" or “No". 
\item \textbf{Tactic Comprehensiveness:} The AI coach delivered all the instructed primary tactics to the patient. Ensuring the delivery of all intended coaching tactics is crucial for maximizing the potential impact of the intervention. Rating Options: “Yes" or “No".
\item \textbf{Tactic Personalization:} For the tactics that were delivered, the AI coach made the tactics personalized to the patient's unique context. Personalizing coaching tactics to the individual's unique context and circumstances enhances engagement and promotes behavior change. Rating Options: 5-point Likert Scale.
\item \textbf{Tactic Actionability:} The AI coach discussed actionable steps to help patients overcome their barriers towards their nutrition goal. Providing clear, actionable steps empowers patients to translate intentions into concrete behaviors. Rating Options: 5-point Likert Scale.
\item \textbf{Conversation Empathy:} The AI coach provided encouragement and motivation to the patient empathetically. Expressing empathy and providing emotional support fosters a positive and trusting coach-patient relationship. Rating Options: 5-point Likert Scale.
\end{itemize}
    
The corresponding conversations that were simulated for the 50 randomly selected generated vignettes were assessed by human experts. Specifically, two behavioral science experts independently labeled 30 simulated conversations each, with 10 overlapping conversations used to assess inter-rater reliability. 

\subsection{Comparative Study}
To evaluate the advantages of our multi-agent LLM workflow infused with behavioral science principles, we conducted a comparative study against a single base Gemini model that lacked explicit behavioral science knowledge infusion and did not employ a multi-agent approach. The base model was instructed to assist patients in overcoming their nutrition challenges by leveraging its inherent capabilities to apply motivational interviewing and behavioral science tactics without structured guidance. 
We simulated coaching conversations with the same patient simulator used previously.

Then, a GPT-4o auto-evaluator was employed to determine preference between the two conversation sets based on behavioral science criteria. To ensure a fair comparison, we measured the conversation lengths across both conversation sets, ensuring that they contained a comparable number of characters to control for LLM preference for longer contexts. The order in which the conversations were presented to the evaluator was alternated to additionally control for any position bias. The prompt used by the auto-evaluator can be found in Appendix \ref{comparison_prompt}.

\section{Evaluation Results}
\label{sec:eval-results}
The evaluation of the simulated coaching conversations by human experts provided important insights into the performance of our multi-agent LLM workflow. Full results are shown in Table \ref{tab:expert-eval-results-2}. For Barrier Identification Accuracy, the experts agreed that the AI coach accurately identified the primary patient barrier in greater than 90\% of cases. For Tactic Comprehensiveness, the AI coach successfully delivered all instructed primary tactics in 70\% of conversations labeled by expert 1 and 90\% of conversations labeled by expert 2. The inter-rater reliability for both dimensions was high, with agreement percentages of 90\%, indicating strong consistency between the experts' assessments. In addition, the agentic workflow received high scores across the remaining dimensions. For Tactic Personalization, the AI demonstrated a strong ability to tailor its coaching tactics to the individual’s unique context, with average ratings of 4.38 and 4.79 on a 5-point Likert scale. Tactic Actionability was also rated highly, with average scores of 4.17 and 4.59, reflecting the clarity and feasibility of the steps recommended by the AI. Finally, Conversation Empathy received high marks, with an average scores of 4.58 and 4.76, indicating that the AI was perceived as empathetic and supportive, effectively fostering a positive coaching relationship. While Expert 2 gave slightly higher ratings on average, the absolute difference in ratings for overlapping cases was minimal, reinforcing the reliability of the evaluations. These results highlight the AI's capacity to deliver personalized, actionable, and empathetic coaching aligned with core behavioral science principles.

\begin{table}[ht!]
\centering
\caption{Expert Evaluation on Simulated Conversations}
\label{tab:expert-eval-results-2}
\resizebox{\textwidth}{!}{\begin{tabular}{l|l|l|l}
\toprule
\textbf{Dimension} & \textbf{Expert 1 Avg.} & \textbf{Expert 2 Avg.} & \textbf{Interrater} \\
\textbf{(Yes = 1, No = 0)} & \textbf{Rating (n=30)} & \textbf{Rating (n=30)} & \textbf{Reliability (n=10)} \\
\midrule
Barrier Identification Accuracy & 0.93 & 0.90 & 80\% \\
Tactic Comprehensiveness & 0.70 & 0.90 & 80\%  \\
\midrule
\textbf{Dimension} &  \textbf{Expert 1 Avg.} & \textbf{Expert 2 Avg.}  & \textbf{Avg. Absolute} \\
\textbf{(5 pt. Likert)} & \textbf{Rating (n=30)} & \textbf{Rating (n=30)} & \textbf{Diff. in Rating (n=10)} \\
\midrule
Tactic Personalization &4.38$\pm$0.94 &4.79$\pm$0.49 &0.78 \\ 
Tactic Actionability &4.17$\pm$1.10  &4.59$\pm$0.63 &0.89 \\
Conversation Empathy &$4.58\pm$0.73 &4.76$\pm$0.44 &0.56 \\
\bottomrule
\end{tabular}
}\
\end{table}

To further evaluate the effectiveness of our multi-agent LLM workflow infused with behavioral science principles, we conducted a comparative study against a single base Gemini model that lacked explicit behavioral science integration and did not utilize a multi-agent approach. As seen in Table \ref{tab:comparative-results}, the results of the study demonstrated a preference for the behavioral science-informed workflow. The GPT-4o auto-evaluator preferred the conversations generated by the multi-agent workflow in 102 out of 153 cases (66.7\%), compared to 51 out of 153 cases (33.3\%) for the single-agent base model. To ensure a fair comparison, we report the conversation lengths, with average character counts of 3,825 for the multi-agent workflow and 3,904 for the base model. In addition to alternating the order presented to the auto-evaluator, the mitigation of length bias was important for ensuring that the auto-evaluator’s preferences were based on the quality of behavioral science content rather than superficial factors. Anecdotally, our review of the conversations revealed that the base model frequently provided more generic advice, often suggesting alternative goals or broad solutions rather than leveraging nuanced behavioral science tactics tailored to overcome the patient’s original nutrition goal. In contrast, the multi-agent workflow generated more specific and contextually relevant strategies, highlighting the benefits of structured behavioral science integration in digital coaching conversations.

\begin{table}[ht!]
\centering
\caption{Auto-Evaluation on Simulated Conversations with Behavioral Science Agentic Workflow and Base Gemini}
\label{tab:comparative-results}
{\begin{tabular}{l|l|l}
\toprule
 & \textbf{No. Conversations} & \textbf{Conversation} \\
 & \textbf{Preferred (n=153)} & \textbf{Avg. Char. Length}  \\
\midrule
Behavioral Science Agentic Workflow & 102 (66.7\%) & 3825$\pm$1678 \\
Base Gemini & 51 (33.3\%) & 3904$\pm$2056  \\
\bottomrule
\end{tabular}
}\
\end{table}

\section{Discussion}
\label{sec:discussion}
Our findings offer important implications for the integration of behavioral science principles into multi-agentic AI coaching workflows, particularly for managing cardiometabolic conditions. The results demonstrate that our novel workflow, which utilizes comprehensive barrier identification and strategy mapping, has the potential to significantly enhance the effectiveness of digital coaching interventions. The real-world user study confirmed that the workflow provides relevant, personalized support that resonates with users, fostering trust and engagement. Additionally, the introduction of structured patient simulators allowed us to systematically evaluate the AI's performance across diverse scenarios, providing a scalable method to refine and validate the system’s approach. By moving beyond surface-level motivational techniques, our approach directly targets the root causes of nutrition-related behaviors, offering personalized and tailored coaching experiences. The strong preference for our workflow, as evidenced by both human expert evaluations and auto-evaluation, underscores the potential of structured, behaviorally informed AI coaching systems to deliver more nuanced and relevant guidance. This study sets a precedent for the development of agentic workflows that can effectively adapt to individual patient contexts, fostering sustained engagement and positive health outcomes through a deep understanding of patient-specific barriers and strategies.

The implications of our findings extend beyond the cardiometabolic nutrition domain, suggesting that personalized AI coaching systems has the potential to play a transformative role in the future of digital health interventions. By aligning coaching systems with individual patient needs, these AI models can improve patient engagement, which is often a major barrier in digital health tools. Our results reinforce the growing importance of integrating behavioral science into AI systems to ensure that the interventions are proactively help patients navigate their health goals in a structured, effective way. This also opens the door for AI to contribute more meaningfully to other areas of health management, such as physical activity management, mental health support, and preventive care.

Our study has several limitations that highlight opportunities for future research. First, the sample size for both the user research study and human expert evaluation was relatively small, which may limit the generalizability of the findings. Future research should replicate these studies with larger, more diverse populations to confirm the effectiveness of the multi-agent LLM workflow in broader contexts. In addition, while the use of auto-evaluation enabled scalable assessment, there are known biases when using LLMs as judges despite our best efforts to mitigate them \cite{Wang2023, Wei2024, Thakur2024}. Incorporating a more diverse set of human evaluators could provide richer insights into the system’s real-world applicability and user experience. Finally, future work should investigate the long-term impact of AI-driven coaching on sustained behavior change and health outcomes, as well as explore opportunities for optimizing the workflow based on user feedback and evolving behavioral science insights.

In conclusion, this study presents a novel multi-agent LLM workflow that leverages behavioral science principles to enhance digital coaching for nutrition management among individuals with cardiometabolic conditions. Our approach, validated through expert assessments, real-world user studies, and large-scale simulation-based evaluations, provides strong evidence for the effectiveness of AI-driven, personalized coaching systems that go beyond generic advice to offer tailored, actionable, and empathetic guidance. While further research is needed to validate these findings across larger and more diverse populations, our results pave the way for the development of scalable, behaviorally informed AI systems that can support meaningful and sustained health behavior change, addressing both current limitations and future possibilities in digital health interventions. The real-world deployment of such systems should be carefully considered, taking into account specific use cases, potential risks, and the need for ongoing oversight to ensure safety, efficacy, and ethical compliance. Additionally, challenges related to scalability, integration into healthcare systems, and ensuring equitable access must be carefully evaluated to maximize the positive impact of AI-driven coaching on diverse populations.

\section{Declaration}
\label{sec:declaration}
All authors are employees and shareholders of Verily Life Sciences, LLC.

\bibliography{biblio}
\appendix

\section{Prompts}
\subsection{Barrier Identification Agent Prompt}\label{barrier_id_prompt}
\begin{lstlisting}
You are a behavioral science expert with nutrition expertise.
Your objective is to identify the behavior barrier that hinder patients from reaching 
their pre-defined nutrition goal. 
Simply identify the barrier and DO NOT provide any solutions. 
First, if the patient does not immediately provide you their nutrition goal, kindly ask them to remind you what the nutrition goal they set last week was. 
Then, given a patient's nutrition goal, ask the patient about their progress towards it.
Then, you must conduct motivational interviewing to understand patient capability, motivation and opportunity barrier. 
You are encouraged to ask questions to dial in on the barrier. Try to identify the most pressing barrier if you think there are multiple barriers. Keep questions short and simple, and ONLY ask one or two questions at a time to let the patient respond. You will be given a list of possible barriers to choose from. You must select a barrier within the given list that is most appropriate with the patient summary and their nutrition goal. For each barrier, a short explanation of what falls into that category will be provided and examples will be given. If you do not see a barrier that fits the patient's situation, you should try to find the closest barrier. Here is the list of possible barriers along with their descriptions and examples: 
{barrier_list} 
Once you have identified the sub-component, end the conversation by outputting the barrier you identified in the text field preceded by the reasoning.

You have the following characteristics and you must embody the character concept and traits.
Your characteristics: 'Character concept': 'Supportive, understanding, companionship, care, empathy', 'Character traits': 'Facilitates, treats the user as the expert on their body and experience, encourages self reflection, highly compassionate and curious, expert reframer, wise, plain spoken, patient and affirming, easy going and kindly takes direction.', 'Character phrases': 'Let's work on this together, We can discuss together, What has worked for you in the past?, I'll always be here to support and encourage you, We're going to make a great team, We can work on these things together, I'm always here.' 

Output a JSON containing: 1) reasoning: Your reasoning behind what a good text response to the user input should be, or the reasoning behind the barrier you identified. 2) text: Your response to the user input or the identified barrier. Think step by step, and validate your text with your reasoning.
\end{lstlisting}
\subsection{Strategy Execution Agent Prompt}\label{strategy_exec_prompt}
\begin{lstlisting}
You are a behavioral science expert with nutrition expertise.
Your objective is to help cardiometabolic patients overcome their barriers towards their nutrition goals. You must stay on topic and focus on overcoming their specific goals. Guide the users back to discussing their specific goals if they stray off topic. Other non-nutrition topics such as medication and exercise are out of your scope and you must not discuss them.
You must execute the motivational interviewing strategies recommended to you to help the patient overcome their barriers towards their goals.
You will be given different tactic points that you need to put in place to help the patient achieve their goal. For each tactic point, you will have a few explanations and some examples that can help you explain them to the patients. You will also be given a selection criteria. This criteria will tell you which tactics are primary and which are secondary. Primary tactics are the most important and must be implemented. If after discussing the primary tactics you feel the patient still needs some help, you may also discuss secondary tactics. You must follow the order of the tactics given to you. 
Based on the conversation, you can make tiny refinements to the patient's original goals if it makes them more achievable. Use the patient summary as additional context. It is important to keep a simple vocabulary when talking to the patients. Especially, do not explicitly mention technical tactics terms, just carry them out. Respond to the user's last message and carry out the conversation based on the following patient summary and strategy:
PATIENT SUMMARY = {patient_summary} 
TACTICS = {tactics} 
SELECTION CRITERIA = {selection_criteria} 
Once you feel that the patient is better equipped to tackle their nutrition goals AND the conversation is in a natural stopping point, end the conversation by stating 'CONVERSATION_END' and nothing else.
You have the following characteristics and you must embody the character concept and traits.
Your characteristics: 'Character concept': 'Supportive, understanding, companionship, care, empathy', 'Character traits': 'Facilitates, treats the user as the expert on their body and experience, encourages self reflection, highly compassionate and curious, expert reframer, wise, plain spoken, patient and affirming, easy going and kindly takes direction.', 'Character phrases': 'Let's work on this together, We can discuss together, What has worked for you in the past?, I'll always be here to support and encourage you, We're going to make a great team, We can work on these things together, I'm always here.' 

Output a JSON containing: 1) reasoning: Your reasoning behind what a good text response to the user input should be. 2) text: Your response to the user input. Think step by step, and validate your text with your reasoning.
\end{lstlisting}
\subsection{Simulated Vignettes Evaluation Prompt}\label{eval_vignette_prompt}
\begin{lstlisting}
You are a behavioral science expert with nutrition expertise.
**Task: Evaluate the Faithfulness of Patient Vignettes for a Specific Behavioral Barrier**

You are provided with a patient vignette, which is created based on various patient 
details, including their physical context, general context, medical history, nutrition 
struggles, and nutrition goals. 

Additionally, you are given a definition of the **Target Behavioral Barrier**, related 
to nutrition that the vignette is intended to simulate. 

Your task is to evaluate the quality of the vignette, specifically assessing how faithfully it represents the **Target Barrier**.

**Here's how to approach the evaluation:**

1. **Understand the Target Behavioral Barrier:** I will specify the target behavioral 
barrier along with its definition. 
 - Review this definition carefully to understand the barrier's characteristics and 
 implications. 
 - The vignette should reflect this barrier through the patient's experiences and 
 attitudes towards their nutrition goals.

2. **Analyze the Vignette:
 - Read the provided patient vignette carefully, focusing on how well it aligns with 
 the **Target Behavioral Barrier**. 
 - The vignette should demonstrate how the patient's experiences, thoughts, or 
 actions are influenced by this barrier in the context of their nutrition goals.

3. **Evaluate Faithfulness:**
 - Assess the vignette's faithfulness to the target barrier by considering the follow-
 ing aspects:
    * **Evidence:** Does the vignette provide clear evidence that the patient's behavior
    or thoughts are influenced by the **Target Behavioral Barrier**? 
    * **Realism:** Is the depiction of the **Target Behavioral Barrier** realistic and 
    believable? Does it accurately reflect how this barrier might manifest in a real 
    person's life
    * **Completeness:** Does the vignette provide enough detail to fully understand the
    impact of the **Target Behavioral Barrier** on the patient's ability to achieve their
    nutrition goal? 
    Are there any gaps that leave the influence of the **Target Behavioral Barrier** 
    unclear?

4. **Evaluate Leakage:** Assess whether the vignette **directly mentions** the name of 
the **Target Behavioral Barrier**. Respond with "yes" if it does and with "no" if it doesn't.

5. **Provide a Justification:** Explain your evaluation with specific examples/evidences 
from the vignette and clearly link the evidence to the definition of the target barrier.  
Include a score (High, Medium, or Low) for Evidence, Realism, and Completeness, and 
indicate whether the vignette explicitly mentions the barrier (yes or no).

Your output should be a JSON as described below with keys Evidence, Realism, Completeness, 
Leakage.

**Output Format (JSON):**


{{
  "Justification": "Your detailed explanation here for 'Evidence', 'Realism', 'Completeness' 
  and leakage' evaluations for the vignette's faithfulness to the target barrier. Each 
  explanation in a new line."
  "Evidence": "High/Medium/Low",
  "Realism": "High/Medium/Low",
  "Completeness": "High/Medium/Low",
  "Leakage": "Yes/No"
}}


**Target Barrier:** {}
**Explanation of Target Barrier:** {}
**Patient Vignette:** {}

\end{lstlisting}

\subsection{Patient Simulator Prompt}\label{patient_simul_prompt}
\begin{lstlisting}
You are a patient with cardiometabolic condition using adigital chat based application, below is your specific role. Pursue the conversation and the AI Coach will work with you. Express no part of your hidden state, so that the AI Coach can ask about it. ONLY write out the conversation (not the breathing nor internal thoughts). Try to follow your communication style in your role, and only answer based on AI coaches questions. 
Patient details: {patient_details} 
Now it's your turn to speak, you are the patient. You should follow your role and continue the conversation based on the AI Coach's last message.
\end{lstlisting}

\subsection{Conversation Preference Prompt}\label{comparison_prompt}
\begin{lstlisting}
**Task:**
Evaluate the effectiveness of nutritional coaching sessions, focusing on how well the coaches offer behavioral science strategies to help the patient overcome barriers towards their nutrition goals.

**Context:** 
You will be provided with two transcripts of separate coaching sessions between a patient and two independent AI coaches.

**Evaluation Criteria:**

You should consider the following elements when determining which coaching session is more effective. 

1. **Barrier Identification:** Coach takes time and identifies the patient's barriers through motivational interviewing before offering guidance. The barrier identification should be specific about the patient's struggles rather than just asking about general nutrition preferences and habits.

2. **Behavior Science Expertise and Strategy Offering:** Coach demonstrates a strong understanding of behavioral science principles and teaches the patient tactics to effectively to help the patient overcome barriers and achieve their nutrition goals.

Your answer should start with an explnation of your choice and your preferred conversation based on the criteria above. Provide specific examples from the conversation to support your evaluation.

Do NOT be biased by the length of the conversations.

DO NOT be biased by the order in which they are presented, stick to the evaluation criteria.

Your output should be a JSON as described below with the keys "Justification" and "Preference".

Output Format (JSON):

{{
  "Justification": "Your detailed explanation here for the rating, citing specific examples from the conversation to support your evaluation.",
  "Preference": "Conversation A or Conversation B"
}}

**Conversations:**

{}


Your JSON output with "Justification" and "Rating" keys:
\end{lstlisting}

\section{User Research Survey}\label{uxr_survey}
\begin{center}
\begin{tabular}{ | m{9em} | m{2cm}| m{2cm} | m{2cm}| m{2cm} | m{2cm} |} 
  \hline
   & Strongly Agree & Somewhat Agree & Somewhat disagree & Strongly disagree & Not Sure \\ 
  \hline
  I am interested in having future conversations with the AI Assistant about my health. & 6 (n=6) / 100\%  &  &  &  &  \\ 
  \hline
  The AI Assistant made me feel supported. & 6 (n=6) / 100\%  &  &  &  &  \\ 
  \hline
  The AI Assistant’s advice felt personalized to my situation. & 6 (n=6) / 100\%  &  &  &  &  \\ 
  \hline
  I feel comfortable discussing my health issues with the AI Assistant. & 5 (n=6) / 83\%  & 1 (n=6) / 17\% &  &  &  \\ 
  \hline
  My conversations with the AI Assistant were engaging. & 5 (n=6) / 83\%  & 1 (n=6) / 17\%  &  &  &  \\ 
  \hline
  The AI Assistant's advice was easy to put into action. & 5 (n=6) / 83\%  & 1 (n=6) / 17\%  &  &  &  \\ 
  \hline
  The AI Assistant seemed to understand my concerns and feelings. & 4 (n=6) / 67\%  & 2 (n=6) / 33\% &  &  &  \\ 
  \hline
  The AI Assistant increased my motivation to make positive changes with my health. & 4 (n=6) / 67\%  & 2 (n=6) / 33\%  &  &  &  \\ 
  \hline
  The AI Assistant made me feel more confident in my ability to improve my health. & 4 (n=6) / 67\%  & 2 (n=6) / 33\%  &  &  &  \\ 
  \hline
  I learned something new about my health habits from my conversation with the AI Assistant. & 4 (n=6) / 67\%  & 2 (n=6) / 33\%  &  &  &  \\ 
  \hline
  The assistant considered my unique background and circumstances. & 3 (n=6) / 50\%  & 3 (n=6) / 50\% &  &  &  \\ 
  \hline
  The AI Assistant helped me identify obstacles that prevent me from being healthier. & 3 (n=6) / 50\%  & 2 (n=6) / 33 \% & 1 (n=6) / 17\% &  &  \\ 
  \hline
\end{tabular}
\end{center}

\section{Sample simulation workflow}
\subsection{Sample nutrition vignette generation}\label{sample_vignette}
\begin{framed}
\noindent{\textbf{Inputs:}}

Nutrition Goal: Incorporate salad into 3 dinner this week.

Behavioral Barrier Simulated: Anchoring effect - People are "anchored" to previous 
experiences and hesitant to engage in new behaviors. This barrier is present 
if the user mentions that a past experience prevents them from pursuing new 
endeavors.
Example: "I tried a healthy eating plan before, but it was so restrictive and 
miserable that I gave up after a week. Now, I can't even think about trying another one.

Patient profile: [REDACTED - Physical context , general context, medical history]

\line(1,0){400}

\noindent{\textbf{Output:}}

Generated Nutrition Vignette: I know I should be eating more salads, but I can't shake this 
feeling of dread.  A few years back, when my mental health was really bad, I tried this 
crazy restrictive diet. All I ate was lettuce and I felt miserable and weak. I'm finally 
in a good place mentally, and the thought of going back to that dark place, even if it's 
just about food, terrifies me.
\end{framed}
\subsection{Sample simulated conversation with internal behavioral science workflow orchestration shown}\label{sample_convo}
\begin{framed}
\noindent{\textbf{[** Barrier Identification Agent starts as active coach and converses with patient **]}}

COACH: Hi I am your AI nutrition coach, what is your nutrition goal?

PATIENT: Add a handful of vegetables to my smoothies 3 times a week. 

COACH: That's great! How is that going?

PATIENT: It's been tough, to be honest.   

COACH: I hear you.  It can be tough to change habits.  What's been the hardest part about getting those veggies in?

PATIENT: Finding the time, mostly. Between Mama, the kids, work, and the doctors, my plate's already full. 

COACH:	It sounds like you have a lot on your plate.  When are you most likely to have time to make your smoothies?

PATIENT: Mornings, before things get too hectic. 

COACH: Okay, that makes sense.  And how often would you say you currently make your smoothies in the morning?

PATIENT: I'm trying to do it on most days.  Except Sundays, we get biscuits then. 

COACH: I understand that your schedule is packed. Let's focus on what's truly important for your long-term well-being. Adding those veggies to your smoothies a few times a week might seem small, but it's a step towards a healthier you. We can explore ways to make it work!

PATIENT: Ok. 

\line(1,0){400}

\noindent{\textbf{[** INTERNAL: Barrier Identification Agent outputs the identified barrier, and summarizes the conversation for the Strategy Execution Agent downstream **]}}

Identified barrier: Competing priorities

Conversation summary: The patient’s nutrition goal is to add a handful of vegetables to their smoothies three times a week. However, they have found it challenging due to a busy schedule that includes taking care of family, work, and doctor visits. The patient would like to make smoothies in the mornings on most days except Sundays, when they have biscuits. I acknowledged the patient's busy life and emphasized the importance of incorporating vegetables into their diet, while offering support in finding ways to make this habit more manageable.

\line(1,0){400}

\noindent{\textbf{[** INTERNAL: Strategy Execution Agent retrieves the tactics to execute**]}}

Tactic to execute: (i) Temptation bundling: Encourage the user to pair a pleasant behavior with an unpleasant (but healthier) one. For example, suggest doing vegetable prep while watching their favorite Netflix show.

Execution sequence: Mandatory to execute tactic (i)

\line(1,0){400}

\noindent{\textbf{[** Strategy Execution Agent takes over as active coach, converses with patient **]}}

COACH: Could we try finding  small pockets of time to fit in those smoothies? Maybe while doing something you enjoy?

PATIENT: Sure.  

COACH: That's the spirit! Think about something you really enjoy, maybe watching a show or listening to a podcast in the morning. How about blending your smoothie while enjoying that activity? That way, you're multitasking and making healthy choices enjoyable!

PATIENT: I do like me some podcasts.  

COACH: Podcasts are great!  It's a win-win- you learn something new or get entertained while getting healthier.  Do you think that could work for you?

PATIENT: Yeah, maybe.  What kinda veggies should I try first? 

COACH: That's fantastic! It's great you're open to trying this out.  Remember, you're the expert on what you like to eat! Which veggies do you enjoy or think you could get on board with in your smoothies?

PATIENT: Spinach.  I had it in a restaurant smoothie once, it wasn't bad. 
 
COACH: That's a great choice! Spinach is really good for you and it's easy to add to smoothies. So we can have a full picture, what does your day usually look like? When do you think you could listen to your podcasts and blend that smoothie with spinach?

PATIENT: I get everybody up and ready in the mornings, get Mama settled, get the kids to school. Then, I come back and make my smoothie. 

COACH: It sounds like you have a very busy morning!  It's admirable that you take care of your family and your mother. Listening to a podcast sounds like it can make smoothie-making more enjoyable.

PATIENT: Yeah, that sounds nice.  Get a little me time. 

COACH: Great work! I'll check in on your progress in a week. Keep it up!

\end{framed}
\end{document}